\documentclass[letterpaper, 10 pt, conference]{ieeeconf}  % Comment this line out if you need a4paper

\IEEEoverridecommandlockouts                              % This command is only needed if 
\usepackage{amsmath}
\usepackage{xcolor}
\usepackage{multirow}
\usepackage{amsfonts}
\usepackage{blindtext}
\usepackage{graphicx}
\usepackage{lineno}
\usepackage[T1]{fontenc}
\usepackage{tabularray}
\title{\LARGE \bf
DenVisCoM: Dense Vision Correspondence Mamba for Efficient and Real-time Optical Flow and Stereo Estimation
}

\author{
Tushar Anand$^{*}$,
Maheswar Bora$^{*}$,
Antitza Dantcheva$^{\dagger}$,
Abhijit Das$^{*}$%
\thanks{$^{*}$Machine Intelligence Group, 
Birla Institute of Technology and Science, Pilani, Hyderabad Campus, India.}%
\thanks{$^{\dagger}$Université Côte d’Azur, Inria, France.}%
\thanks{Email: \texttt{abhijit.ads@hyderabad.bits-pilani.ac.in}}%
}

\begin{document}

\maketitle
\thispagestyle{empty}
\pagestyle{empty}

\begin{abstract}
In this work, we propose a novel Mamba block DenVisCoM, as well as a novel hybrid architecture specifically tailored for accurate and real-time estimation of optical flow and disparity estimation. Given that such multi-view geometry and motion tasks are fundamentally related, we propose a unified architecture to tackle them jointly. Specifically, the proposed hybrid architecture is based on DenVisCoM and a Transformer-based attention block that efficiently addresses real-time inference, memory footprint, and accuracy for at the same time for joint estimation of motion and 3D dense perception tasks. We extensively analyze the benchmark trade-off of accuracy and real-time processing on a large number of datasets. Our experimental results and related analysis suggest that our proposed model can accurately estimate optical flow and disparity estimation in real time. All models and associated code are available at https://github.com/vimstereo/DenVisCoM.

\end{abstract}    
\section{Introduction}
\label{sec:intro}

Estimating optical flow and stereo disparity are fundamental challenges in computer vision, crucial for applications ranging from robotics to autonomous driving~\cite{hamzah2016literature}.  Early approaches, often relying on patch matching and assuming perfect lens rectification~\cite{liang2018learning}, have been superseded by deep learning methods.  While CNNs~\cite{hamid2022stereo, lipson2021raft} and transformers~\cite{xu2023unifying} offer improved accuracy, they face a significant trade-off between computational efficiency and performance, hindering real-time deployment, particularly on resource-constrained platforms.  Transformers, known for their ability to capture long-range dependencies, suffer from quadratic computational complexity, making them expensive to train and deploy.

State Space Models (SSMs), particularly Mamba~\cite{gu2023mamba} and its successors~\cite{dao2024transformersssmsgeneralizedmodels}, present a promising alternative.  SSMs achieve linear complexity and efficient parallel training, potentially resolving the accuracy-efficiency dilemma.  However, directly applying Mamba, originally designed for sequential data, to vision tasks is non-trivial.  Adaptations like Vision Mamba (ViM)~\cite{zhu2024vision}, MambaVision~\cite{mambavision}, VMamba~\cite{liu2024vmambavisualstatespace}, and NC-SSD~\cite{dao2024transformersssmsgeneralizedmodels} address this by incorporating positional awareness and modified scanning strategies.  These primarily focus on single-image tasks.  Emerging multi-modal Mamba models~\cite{li2024mambadfusemambabaseddualphasemodel, li2024cfmwcrossmodalityfusionmamba, dong2024fusionmambacrossmodalityobjectdetection} exist, but they do not directly address the core requirement of visual dense correspondence matching between image pairs, crucial for optical flow and stereo disparity. In this direction, very recently, ViM-disparity was proposed by \cite{bora2025vim}, employing ViM for accurate and real-time disparity map estimation. Although extensive. 
However, dense correspondence tasks demand an explicit comparison of corresponding features, necessitating feature representations that encode both global position and local relationships.  Existing approaches lack this nuanced correspondence handling.

Hence, in this paper, we introduce a novel hybrid architecture, optimized for optical flow and stereo disparity, that fundamentally rethinks the Mamba block.  Our central innovation is a visual correspondent mechanism that fuses image pair features patch-wise within the Mamba sequence transformation.  To enable robust fusion and capture essential relationships, we integrate self- and cross-attention, addressing Mamba's inherent lack of cross-correspondence.  This builds upon the demonstrated success of hybrid SSM-transformer architectures in vision. 

Our key contributions are:

\begin{itemize}
    \item A novel hybrid SSM architecture combining Mamba with self- and cross-attention, specifically tailored for accurate and efficient optical flow and stereo disparity estimation.
    \item A redesigned Mamba block that facilitates joint learning of image pair features via a visual correspondence mechanism within the sequence transformation.
%    \item Extensive benchmarking, demonstrating superior performance in terms of inference speed, accuracy, and memory efficiency, highlighting suitability for real-time, resource-constrained scenarios.
\end{itemize}

\section{Related Work}
\label{sec:related_work}

\subsection{Mamba Architectures}

Mamba~\cite{gu2023mamba, gu2024mambalineartimesequencemodeling} emerged as a state-space model (SSM) offering a computationally efficient alternative to transformers, reducing complexity from quadratic to linear while maintaining competitive performance in language modeling tasks.  This success prompted efforts to adapt Mamba to vision.  Key challenges included Mamba's unidirectional nature and lack of inherent positional awareness.

Vision Mamba (Vim)~\cite{zhu2024vision} addressed these by incorporating bidirectional SSMs and position embeddings.  However, bidirectionality can introduce latency.  MambaVision~\cite{mambavision} adopted a hybrid approach, combining modified Mamba blocks with transformer blocks.  VMamba~\cite{liu2024vmambavisualstatespace} introduced a cross-scanning mechanism to improve information flow.  Mamba-2~\cite{dao2024transformersssmsgeneralizedmodels}, based on the state space duality (SSD) framework, refined the selective SSM. VSSD~\cite{shi2024vssdvisionmambanoncausal} further enhanced this with a non-causal block (NC-SSD), maintaining global receptive fields and linear complexity while improving training and inference.

Beyond single-image tasks, Mamba has been explored for multi-modal applications, including medical image fusion (MambaDFuse)~\cite{li2024mambadfusemambabaseddualphasemodel}, multi-instance learning~\cite{li2024cfmwcrossmodalityfusionmamba}, object detection~\cite{dong2024fusionmambacrossmodalityobjectdetection}, and the fusion of multispectral and RGB images~\cite{zhou2024dmm}. However, these existing multi-modal Mamba models do not directly address the specific requirements of optical flow and stereo disparity estimation, particularly the need for a robust mechanism to establish and leverage visual dense correspondence between image pairs. Our work fills this gap by introducing a novel Mamba block architecture specifically designed for this purpose.
\section{Proposed Methodology}
\label{sec:proposed_methodology}

\subsection{Preliminaries}

Mamba is based on SSMs that map a 1-D function $x(t) \in \mathbb{R} \rightarrow y(t)\in \mathbb{R}$ through a hidden state $h(t)\in \mathbb{R}^N$. It formulates $A \in \mathbb{R}^{N \times N}$ as the evolution parameter, and $B \in \mathbb{R}^{N\times1}$ and $C\in \mathbb{R}^{1\times N}$ as projection parameters:
\begin{align}
h(t) = A h(t - 1) + B x(t), \quad y(t) = Ch(t)
\end{align}

%Mamba2\cite{dao2024transformersssmsgeneralizedmodels} recently simplified the matrix A into a scaler. When A\textsubscript{i} is reduced to a scalar, the linear formula is as follows: 

Mamba are the discrete versions of the continuous
system or SSMs, that include a timescale parameter $\Delta$ to transform the continuous parameters $A$ and $B$ to discrete parameters $\overline{A}$ and $\overline{B}$. The commonly used technique for this transformation is zero-order hold (ZOH). After the discretization of $\overline{A}$ and $\overline{B}$, the discretized version of the above equation using a step size of $\Delta$ is:
\begin{align}
h_t = \overline{A} h_{t - 1} + \overline{B} x_t, \quad y_t = C h_t.
\end{align}

At last, the models compute output through a global convolution
\begin{align}
\overline{K} = (C\overline{B},C\overline{A}\overline{B},  \dots,C\overline{A}^{M-1} \overline{B}), \quad y = x*\overline{K},
\end{align}
where $M$ is the length of the input sequence $x$, and $\overline{K}\in \mathbb{R}^{M}$ denotes a structured convolutional kernel.  To extend Mamba for the vision task, MambaVision modified the Mamba block. Assuming an input $X(t) \in \mathbb{R}^{T\times C}$ with sequence length $T$ with embedding dimension $C$, the following equations describe the interaction between the Mixer and MLP blocks in MambaVision, where feature normalisation is followed by a Mixer and MLP block to process and combine the input representations:
\begin{align}
\hat{X}^n &= \operatorname{Mixer}(\operatorname{Norm}(X^{n-1})) + X^{n-1} \\
X^n &= \operatorname{MLP}(\operatorname{Norm}(\hat{X}^n)) + \hat{X}^n.
\end{align}

% Second set of equations - MambaVision Mixer
The MambaVision Mixer operation combines sequential and spatial information using two branches, Scan \textit{i.e.,} denoted as SSM and Convolution, and concatenates the results $X_{\text{out}}$.
\begin{align}
X_1 &= \operatorname{SSM}(\sigma(\operatorname{Conv}(\operatorname{Linear}(C, C/2)(X_{\text{in}})))), \\
X_2 &= \sigma(\operatorname{Conv}(\operatorname{Linear}(C, C/2)(X_{\text{in}}))), \\
X_{\text{out}} &= \operatorname{Linear}(C/2, C)(\operatorname{Concat}(X_1, X_2)).
\end{align}
Each output is projected to be half the size of the original embedding dimension to maintain a similar number of parameters to the original block.

\subsection{Proposed DenVisCoM Block}
\begin{figure}[t]
  \centering
   \includegraphics[width=\columnwidth]{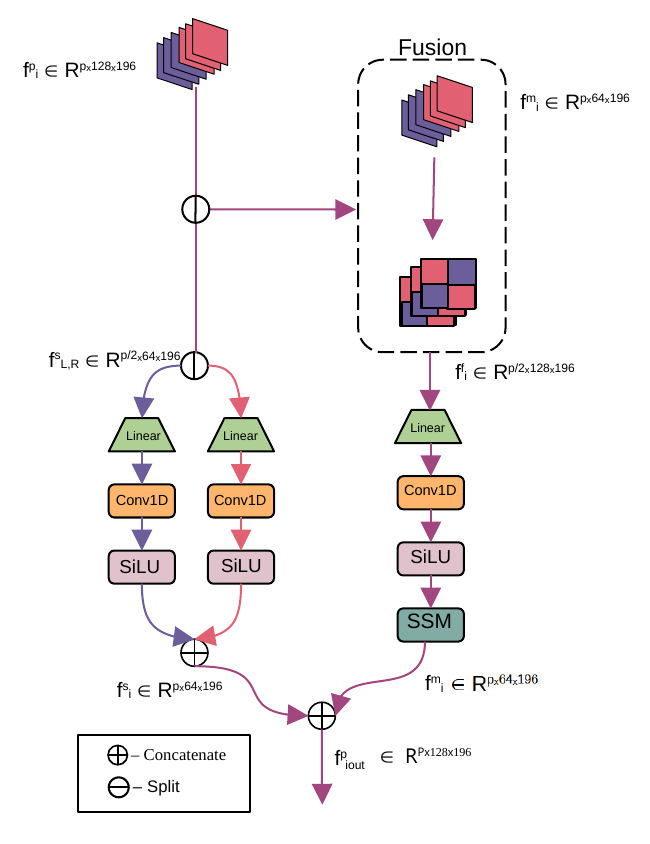}
\caption{The proposed modified DenVisCoM block for sequential joint modeling in the sequence transformation block of Mamba.}
%\label{fig:model_architecture}
\label{fig_proposedMamaba}
\end{figure}

The proposed Mamba block, DenVisCoM, aims to model the dense vision correspondence between a pair of features by jointly learning the pair of input images in the sequence transformation pipeline (See Figure \ref{fig_proposedMamaba}). To achieve this, we introduce two symmetrical convolution branches for each input feature and one joint Scan branch, which takes block-wise fused features (as in the fusion block of Figure \ref{fig_proposedMamaba}). Each convolution branch consists of a linear projection layer, a Conv1D layer, and a SiLU activation function.  The purpose of these symmetric paths is to independently compensate for any information loss in both the left and right images caused by the SSM branch due to its sequential nature.
\begin{align}
    X_{L} &= \sigma(\operatorname{Conv1D}(\operatorname{Linear}(f_{iL}^{s}))) \\
    X_{R} &= \sigma(\operatorname{Conv1D}(\operatorname{Linear}(f_{iR}^{s})))
\end{align}

The Scan branch, \textit{i.e.,} the SSM branch, consists of a linear projection layer, a Conv1D layer, a SiLU activation function, and an SSM for handling long-range dependencies jointly for both inputs while preserving locality with convolutional layers.
\begin{align}
    X_{\text{SSM}} &= \operatorname{SSM}(\sigma(\operatorname{Conv1D}(\operatorname{Linear}(\operatorname{Fusion}(f_{i}^{m})))))
\end{align}
Further, the output of each convolution branch is concatenated, followed by further concatenation with the Scan branch to obtain the output:
\begin{align}
     X_{\text{out}} &= \operatorname{Concat}(X_{\text{SSM}}, \operatorname{Concat}(X_{L},X_R)).
\end{align}

\subsection{Proposed Architecture}
\begin{figure*}[t]
  \centering
   \includegraphics[width=0.7\textwidth]{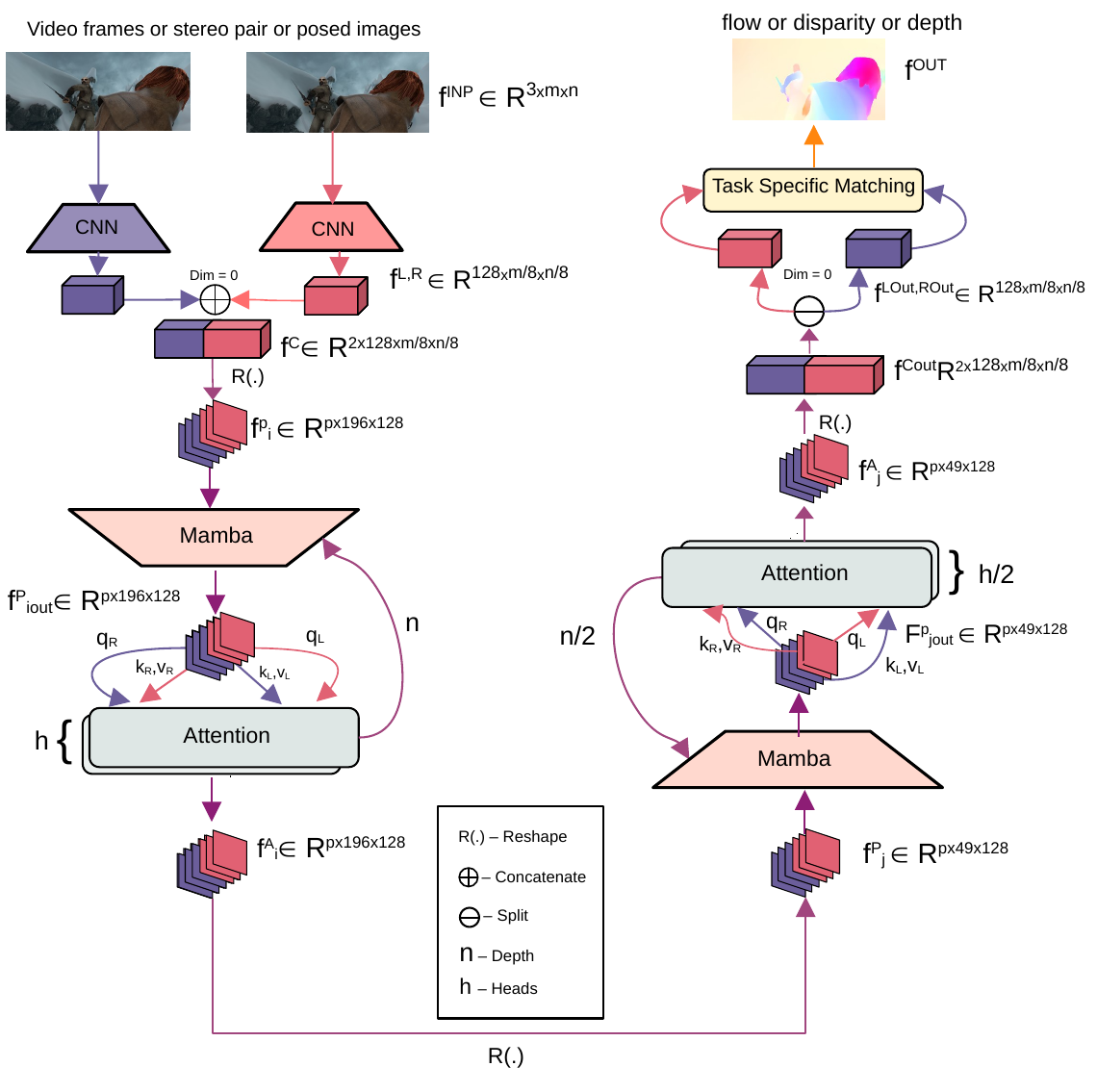}
   \caption{Overview of the proposed hybrid model.}
   \label{fig:proposedarch}
\end{figure*}

To adapt and efficiently extract features via Mamba, the proposed model passes the input images via CNN and then to the Mamba block in the form of patches (See Figure \ref{fig:proposedarch}). Specifically, the proposed model is a hybrid SSM-based approach, which consists of SSM and Transformer-based attention blocks, as in MambaVision.  Precisely, in our proposed architecture, feature normalization is followed by SSM and attention blocks in order to process and combine the dense correspondence of input representations of a pair of input images. We proceed to explain the architecture in detail.

The two left and right (for stereo disparity estimation task) or consecutive frames (for flow) as images ($f_{INPL}, f_{INPR} \in \mathbb{R}^{3 \times m \times n}$) are passed through two separate ResNet18 \cite{he2015deepresiduallearningimage}-based CNN encoders to extract their corresponding 8$\times$ downsampled feature representations ($f^{L}, f^{R} \in \mathbb{R}^{128 \times \frac{m}{8} \times \frac{n}{8}}$), where $m$ and $n$ represent the spatial dimensions.
\begin{align}
 f^{L} = \operatorname{CNN}(f^{INPL}), \quad f^{R} = \operatorname{CNN}(f^{INPR})
\end{align}

The extracted features are then concatenated, and positional embeddings are added to form a combined feature ($f^{C} \in \mathbb{R}^{2 \times 128 \times \frac{m}{8} \times \frac{n}{8}}$).  After concatenation, the concatenated features are reshaped into patches ($f_{i}^{p} \in \mathbb{R}^{p \times 196 \times 128}$) for better local feature extraction. Here, $p$ refers to the number of patches and 196 to the patch size of 14x14. $f^{C} \in \mathbb{R}^{2 \times 128 \times \frac{m}{8} \times \frac{n}{8}}$.
\begin{align}
 f^{L} &= f^{INPL}+P, \quad f^{R} = f^{INPR}+P \\
 f^{C} &= \operatorname{Concat}(f^{L},f^{R}) \\
 f_{i}^{p} &= \operatorname{Patch}(f^{C})
\end{align}

The patches are then passed into the DenVisCoM block with $f_{i}^{p}$ as input (refer to Figure~\ref{fig_proposedMamaba}).  $f_{i}^{p}$ is reshaped to $\mathbb{R}^{p \times 128 \times 196}$ and split along the embedding dimension into $f_{i}^{m}, f_{i}^{s} \in \mathbb{R}^{p \times 64 \times 196}$.

Then, $f_{i}^{s}$ is further split along the patch dimension into the left and right embeddings ($f_{iL}^{s}, f_{iR}^{s} \in \mathbb{R}^{p/2 \times 64 \times 196}$). $f_{i}^{m}$ is split into left and right patches and concatenated along the embedding dimension. This fusion along the embedding dimension allows the corresponding patches of the left and right images to be passed through the SSM branch simultaneously for joint learning, while $f_{iL}^{s}$ and $f_{iR}^{s}$ pass through the each symmetric convolution branches.

After passing through the symmetric branches, $f_{iL}^{s}$ and $f_{iR}^{s}$ are concatenated along the patch dimension back into $f_{i}^{s}$. The output of the SSM branch ($f_{i}^{m} \in \mathbb{R}^{p/2 \times 128 \times 196}$) is unfused to the original tensor $f_{i}^{m} \in \mathbb{R}^{p/2 \times 128 \times 196}$ for further propagation in the network. Finally, $f_{i}^{m}$ and $f_{i}^{s}$ are concatenated along the embedding dimension and reshaped into $f_{i}^{p}$ and passed through a linear projection layer to get the output of the DenVisCoM as $f_{iout}^{p}$.

To further enhance joint learning between the pair of input images, we include both self-attention and cross-attention mechanisms. Multi-head attention improves the diversity of attention heads, allowing the model to capture multiple aspects of the feature space. To reduce computational complexity, we scale the number of attention heads across stages while maintaining a constant number of parameters. Each attention block includes two components: self-attention within each sequence and cross-attention between corresponding image sequences. The self-attention is applied to individual feature maps, followed by cross-attention to enable joint learning between the image pair.

Inside the attention block, $f_{iout}^{p}$ is split into $f_{L}^{a}, f_{R}^{a} \in \mathbb{R}^{p/2 \times 196 \times 128}$. During self-attention, $f_{L}^{a}$ and $f_{R}^{a}$ are the key, query, and value for the left and right images, respectively. During cross-attention, first $f_{R}^{a}$ is the query and $f_{L}^{a}$ acts as the key and value, then $f_{L}^{a}$ becomes the query and $f_{R}^{a}$ acts as the key and value.
\begin{align}
f_{L}^{a} &= \operatorname{SelfAttn}(f_{L}^{a}) \\
f_{R}^{a} &= \operatorname{SelfAttn}(f_{R}^{a}) \\
f_{L}^{a} &= \operatorname{CrossAttn}(Q=f_{R}^{a}, K=f_{L}^{a}, V=f_{L}^{a}) \\
f_{R}^{a} &= \operatorname{CrossAttn}(Q=f_{L}^{a}, K=f_{R}^{a}, V=f_{R}^{a})
\end{align}

Thus, each image feature map attends to both itself and the other image's feature map, enabling richer multi-view representations and improving the overall model's ability to learn visual correspondences between images. After the attention block, $f_{L}^{a}$ and $f_{R}^{a}$ are concatenated along the patch dimension to reconstruct $f_{ia}^{p}$.

The architecture consists of two sequential Mamba and attention blocks. The Mamba and attention blocks are interleaved for improved feature representation and handling of long-range feature dependencies. The interleaving is done for $n$ times, representing the depth with $h$ attention heads. In the first set of Mamba and attention blocks, after the first $n$ passes, the patch size is reduced to 7, forming a tensor $f_{j}^{p} \in \mathbb{R}^{p \times 49 \times 128}$.  In the first set of Mamba and attention blocks, the process is repeated for $n/2$ passes with $2h$ attention heads. Finally, $f_{ja}^{p}$ is reconstructed into $f^{cOut}$ and split along the batch dimension into $f^{LOut}$ and $f^{ROut}$, which are then used for task-specific matching.

\subsection{Task Specific Matching}

For task-specific matching, we take a pair of images, which can be video frames, stereo images or posed images as I\textsubscript{1} and I\textsubscript{2}. A parameter-free task-specific matching layer is used for optical flow and stereo matching. The matching layer takes the 8 $\times$ downsampled feature which have been processed through the Fusion Mamba, f\textsubscript{1},f\textsubscript{2} $\in \mathbb{R}$\textsuperscript{D$\times$H$\times$W}, here H and W denote the height and width, respectively and D refers to the feature dimension.

To compute the optical flow, global matching computes the correspondence of every location in f\textsubscript{1} with every location in f\textsubscript{2} by using matrix multiplication. To avoid very large values from the dot product,$\frac{1}{\sqrt{D}}$ is used as a normalization factor. Further, a softmax layer is added to obtain the dense correspondences. This step normalizes the last two dimensions of C\textsubscript{flow} and produces a distribution of each location in f\textsubscript{1} with respect to each location in f\textsubscript{2}. The correspondence $\hat{G}$\textsubscript{2D} is obtained using the flow with 2D coordinates of pixel grid  G\textsubscript{2D} dimension. Finally, the optical flow V\textsubscript{flow} is calculated by computing the difference between $\hat{G}$\textsubscript{2D} and G\textsubscript{2D}.

The aim of stereo matching is to find disparity along the horizontal scan line. To find the 1D correspondence, it can be treated as a special case of 2D global matching. The last dimension of C\textsubscript{disp} is normalized and a matching distribution. As the correspondence of each pixel in the f\textsubscript{1} is located to the left of the reference pixel, the upper triangle of the W × W slices of image is masked to avoid unnecessary matches. Subsequently, the 1D correspondence is obtained by computing M\textsubscript{disp} with all horizontal locations P $\in$ $\mathbb{R}$\textsuperscript{W}. The final positive disparity is obtained using the difference of corresponding coordinates in $\hat{G}\textsubscript{1D}$ with 1d pixel grid G\textsubscript{1D}.

\section{Experiments and Analysis}
\label{ref:exp_and_analysis}

We train the flow models on the Sceneflow (Flyingthings, Monka and driving) datasets as per the MemFlow protocol[7] for 100k steps, with a batch size of 8, learning rate of 10\textsuperscript{-3} and AdamW optimizer. and performed zero-shot testing on KITTI15(test set) and Sintel. The primary metric used is end-point-error (EPE), the L2 distance between estimated and ground
truth flow vectors. Further, EPE is also reported for motion
ranges s0--10, s10--40, and s40+. We also employed the F1-all
measure (F1A), which indicates the percentage of predicted
flow vectors that deviate significantly from the ground truth
flow, exceeding a certain threshold (usually 3 pixels) across
all pixels in an image. In addition, the frame per second
(FPS), memory required (M) are also reported. For disparity, a similar training protocol was followed and zero-shot testing was performed on KITTI15(test set), VKITTI1 and Sintel. We report these results in \ref{tab_disparity_results} for the model trained from scratch. We evaluate performance using common metrics like EPE, D1, FPS, and memory usage. EPE represents the average L1 distance between predicted and ground truth disparity, whereas D1 indicates the percentage of outliers.

We used 4x RTX A6000 (48GB) with AMD EPYC 9124 16-Core Processor for our training. For FPS and memory benchmarking we utilize a single RTX A6000.

\subsection{Optical Flow}

%\subsubsection{Implementation details and Metrics}
%We train the flow models on KITTI15 and Sintel for 100k steps, with a batch size of 4. The primary metric used is end-point-error (EPE), the \(l_2\) distance between estimated and ground truth flow vectors. EPE is reported for motion ranges \(s_{0-10}\), \(s_{10-40}\), and \(s_{40+}\). In addition, FPS and memory are measured to evaluate real-time performance.

\begin{table*}[htb!]
\centering
\caption{Results on KITTI15 for flow task.}
\vspace{-2.5mm}
\label{flow_results}
\scriptsize
\begin{tblr}{
  width = \linewidth,
  colspec = {Q[220]Q[79]Q[63]Q[83]Q[98]Q[105]Q[99]Q[180]},
  vlines,
  hline{1-2,3,4,5,6,7,8,9} = {-}{},
  hline{3,5,7} = {2-7}{}
}
\textbf{Method} & \textbf{EPE} & \textbf{F1-all} & \textbf{\textit{S\textsubscript{0-10}}} & \textbf{\textit{S\textsubscript{10-40}}} & \textbf{S\textsubscript{40+}} & \textbf{FPS} & \textbf{Memory} \\
RAFT\cite{lipson2021raft}('20)                            &  2.45 & 7.9 & 0.43 & 1.18 & 5.7 & 11.7 & \textbf{180.51} \\
Unimatch~\cite{xu2023unifying}('23)                        &  2.25 & 7.2 & 0.48 & 1.1 & 5.12 & 33.88 & 236.58 \\
MemFlow\cite{dong2024memflowopticalflowestimation}('24)    & 3.38 & 12.8 & 0.46 &1.09& 5.3 & 35.27 & 241.57 \\ 
HD3\cite{hd3}&1.31&6.5&-&-&-&-&-\\
PerceiverIO\cite{jaegle2022perceiveriogeneralarchitecture}&4.98&5.4&-&-&-&-&-\\
ViMDisparity\cite{10888223}&2.73&7.41&0.51&1.13&4.94&32.98&238.54\\
% \textbf{VSSD\cite{shi2024vssdvisionmambanoncausal}('24)}            & 5.581 & 14.14 & 0.207 & 0.95 & 15.58 & 24.76 & 208.41 \\ 
% \textbf{MambaVision\cite{mambavision}('24)}            & 2.04 & 6.95 & 0.42 & 0.87 & 4.79 & 37.83 & 228.17 \\ 
% \textbf{Proposed w/o Fusion}                                        & 0.942 & 2.88 & \textbf{0.15} & 0.42 & 2.25 & \textbf{39.88} & 244.89 \\
% \textbf{Proposed}                                                   & \textbf{0.862} & \textbf{2.52} & \textbf{0.15} & \textbf{0.41} & \textbf{2.00} & 39.39 & 251.03 \\
\textbf{DenVisCoM (Ours)} & \textbf{1.34} & \textbf{2.52} & \textbf{0.28} & \textbf{0.74} & \textbf{3.15} & \textbf{39.88} & 244.89

\end{tblr}
\end{table*}

\subsubsection{Results and Analysis}
The performance of our proposed model on the KITTI dataset, as shown in Table~\ref{flow_results}, demonstrates significant improvements over RAFT, MemFlow and Unimatch. Our model achieves the lowest End-Point Error (EPE) of 1.34 outperforming MemFlow (3.38), RAFT (2.45) and Unimatch (2.25). Additionally, it records the lowest (F1-all) at 2.52, compared to 7.9 and 7.2 for RAFT and Unimatch, respectively, indicating superior robustness in handling challenging cases of optical flow estimation. In terms of motion magnitude, our model excels particularly in the mid-range (S\textsubscript{10-40}) and large motion categories (S\textsubscript{40+}), with EPE values of 0.41 and 2.00, respectively, outperforming RAFT (1.18, 5.7) and Unimatch (1.1, 5.12). The improvement across all motion ranges underscores the effectiveness of our approach in handling complex flow estimation scenarios. We have also compared our proposed method against a non-causal Mamba implementation VSSD\cite{shi2024vssdvisionmambanoncausal} and MambaVision\cite{mambavision}(see \ref{tab:7}), and our proposed method significantly outperforms both.  Overall, the proposed architecture consistently demonstrates superior accuracy, particularly for mid-range and large motion estimation on KITTI, and shows competitive performance. The combination of cross-attention and proposed Mamba blocks significantly enhances the model's ability to handle diverse motion magnitudes, although further optimization may be necessary for large displacements in more complex datasets.

The proposed model shows a significantly improved FPS compared to others, with memory requirements comparable to RAFT and Unimatch. This demonstrates that the proposed method is more effective than quadratic attention for long video sequences while maintaining accuracy.

The performance of the flow task of proposed model was evaluated on the Sintel dataset, with results indicating significant improvements over existing methods such as Unimatch and RAFT, as detailed in Table \ref{tab:1}. Specifically, our model achieved the lowest unmatched error on the Sintel (Final) dataset, with a score of 10.670, outperforming Unimatch (12.74) and FlowFormer (11.37). For the Sintel (Clean) dataset, our model ranked third in both matched and unmatched error, achieving 0.44 and 7.903, respectively, compared to Unimatch, which recorded 0.34 for matched error and 6.68 for unmatched error. Furthermore, our model also secured third place in matched error on the Sintel (Final) dataset.

These results highlight the effectiveness of proposed approach, demonstrating its ability to deliver competitive optical flow estimates with relatively minimal training effort.

\begin{table}[!htp]\centering
\caption{Result of flow task on Sintel. † represents the method that uses the last frame’s flow prediction as initialization for subsequent refinement, while other methods all use two frames only
}
\label{tab:1}
\scriptsize
\begin{tabular}{|l|r|r|r|r|r|}\hline

\multirow{2}{*}{Methods} &\multicolumn{2}{c}{Sintel Clean} \vline &\multicolumn{2}{c}{Sintel Final}\vline \\\cline{2-5}
&matched &unmatched &matched &unmatched \\ \hline
FlowNet2 &1.56 &25.4 &2.75 &30.11 \\\hline
PWC-Net+ &1.41 &20.12 &2.25 &23.7 \\\hline
HD3 &1.62 &30.63 &2.17 &24.99 \\\hline
VCN &1.11 &16.68 &2.22 &22.24 \\\hline
DICL &0.97 &16.24 &1.66 &19.44 \\\hline
%RAFT &- &- &- &- \\\hline
RAFT†\cite{lipson2021raft} &0.62 &9.65 &1.41 &14.68 \\\hline
GMA† &0.58 &7.96 &1.24 &12.5 \\\hline
DIP† &0.52 &8.92 &1.28 &15.49 \\\hline
AGFlow† &0.56 &8.54 &1.22 &12.64 \\\hline
CRAFT† &0.61 &8.2 &1.16 &12.64 \\\hline
FlowFormer &0.41 &7.63 &\textbf{0.99} &11.37 \\\hline
GMFlowNet &0.52 &8.49 &1.27 &13.88 \\\hline
GMFlow\cite{xu2022gmflow} &0.65 &10.56 &1.32 &15.8 \\  \hline
Unimatch\cite{xu2023unifying} &\textbf{0.34} &\textbf{6.68} &1.1 &12.74 \\\hline
Proposed & 0.44 &7.903 &1.173 &\textbf{10.67} \\\hline

\end{tabular}
\end{table}

%\subsection{Cross Task Transfer}
\subsubsection{Ablation on Optical Flow}
The ablation study evaluates the performance of different mamba implementations like Mamba2 and VSSD against our proposed model on the KITTI15, Sintel and Flying-Chairs datasets for optical flow. The study considers two key factors: mamba block implementation and the effect of fusion on the performance of the proposed model. (See Table~\ref{2}). The study considers 4 architectural configurations, proposed where the corresponding left and right patches are being passed through the Mamba1 block simultaneously. The second architecture is a modification of our proposed architecture, here, the patches are passed sequentially. The third architecture replaces our implementation with Mamba2, and the fourth architecture replaces our implementation with VSSD.

{\textbf{KITTI15 Results:}} The proposed configuration demonstrates the best performance with an End-Point Error (EPE) of 0.87 and a relatively low error in handling large motions (s$_{40+}$: 2.24). This showcases the effectiveness of the simultaneous passing of corresponding patches through Mamba1 in capturing motion details, especially in complex flow scenarios. Both proposed without fusion and Mamba2 show competitive performance, with proposed without fusion achieving the lowest large-motion error (s$_{40+}$: 2.13). However, it still shows a higher EPE of 0.93 compared to proposed. The model used here was first trained on Flying-chairs for 50k epochs with a batch size of 4 and subsequently finetuned on Kitti15(train set) for 20k steps with batchsize of 4.

\textbf{Sintel Results:} For the Sintel dataset, our proposed model achieves the best results in terms of overall EPE (1.92). Proposed model without fusion and Mamba2 also show good results with an EPE of 2.09 and 2.14 respectively. VSSD performs slightly worse at an EPE of 2.68. The model used here was first trained on Flying-chairs for 50k epochs with a batch size of 4 and subsequently finetuned on Sintel for 20k steps with batchsize of 4.

%{Flying-Chairs Results:} On the more challenging flying-chairs dataset, the proposed model continues to demonstrate the best performance with an EPE of 2.84, all other configurations. The proposed model without the fusion approach also excels with an EPE of 3.03. The Mamba2 configuration, while still competitive, records a higher EPE of 3.28, while VSSD struggles with an EPE of 3.99. The model used here was first trained on Flying-chairs for 50k epochs with a batch size of 4.

The ablation study reveals that the proposed model with simultaneous passing of left and right patches through Mamba1 block achieves the best performance across datasets, particularly excelling in small and large motion categories. We note that while Mamba2 performs well on KITTI15 and Sintel.

% \begin{table}[!ht]
%     \centering
%     \caption{Ablation study on finetuning for flow task on KITTI15.}
%     \begin{tabular}{|c|c|c|c|c|c|}
%     \hline
% Method & EPE & F1-all & S_{0-10} & S_{10-40} & S_{40+} \\ \hline
%     Proposed  & 0.87  & 2.88 &0.17 &0.48 &2.24\\ \hline
%     Proposed w/o Fusion   & 0.93& 3.05 &0.17 &0.47&2.13     \\ \hline
%     Mamba  & 1.077  & 3.42 &0.19&0.52&2.49  \\ \hline
%     VSSD   & 1.74 &5.74&0.33&0.85&3.89     \\ \hline
%     \end{tabular}
% \label{2}
% \end{table}
\begin{table}[!ht]
\centering
\caption{Ablation study for flow task on KITTI15 and Sintel\cite{geiger2013vision}.}
\resizebox{\columnwidth}{!}{%
\begin{tabular}{|c|c|c|c|c|c|c|}
\hline
\multirow{2}{*}{Methods} &\multicolumn{5}{c}{KITTI15} \vline &\multicolumn{1}{c}{Sintel}\vline \\\cline{2-7}
 & EPE & F1-all & $S_{0-10}$ & $S_{10-40}$ & $S_{40+}$ & EPE\\ \hline
Proposed & \textbf{0.87} & \textbf{2.88} & \textbf{0.17} & 0.48 & 2.24 &\textbf{1.92}\\ \hline
Proposed w/o Fusion & 0.93 & 3.05 & \textbf{0.17} & 0.\textbf{47} & \textbf{2.13}&2.09 \\ \hline
Mamba2 & 1.07 & 3.42 & 0.19 & 0.52 & 2.49&2.14 \\ \hline
VSSD\cite{shi2024vssdvisionmambanoncausal} & 1.74 & 5.74 & 0.33 & 0.85 & 3.89&2.68 \\ \hline
\end{tabular}%
}
\label{2}
\end{table}
\vspace{-4mm}
\begin{table}[ht]
\scriptsize
\centering
\caption{Ablation study for disparity task.}
\label{disp_abalation}
\begin{tblr}{
  width = \linewidth,
  colspec = {Q[250]Q[130]Q[230]Q[230]Q[230]Q[190]Q[230]Q[230]},
  cell{2}{1} = {r=2}{},
  cell{4}{1} = {r=2}{},
  cell{6}{1} = {r=2}{},
  %vlines,
  hline{1-2,10} = {1-8}{},
  hline{3,5,7} = {2-8}{},
  hline{4,6,8} = {1-8}{},
  vline{1,2,3,4,5,6,7,8,9} = {-}{}
}
        &                                   & Proposed  & Proposed w/o Self Attn & Proposed w/o Cross Att.  & Proposed w/o Attention\\
\textbf{Kitti15}\cite{geiger2013vision}     & EPE & \textbf{0.27}      & 0.3        & 0.32     & 0.39 \\
                                            & D1  & \textbf{0.005}     & 0.006          & 0.007     & 0.007\\
\textbf{Vkitti}\cite{cabon2020virtual}       & EPE &\textbf{0.18}      & 0.24        & 0.26      & 0.31  \\
                                             & D1  & \textbf{0.044}     & 0.045          & 0.047     & 0.047           \\
\textbf{Sintel}\cite{butler2012naturalistic} & EPE & \textbf{0.51}      & 0.79          & 0.857      & 0.901            \\
                                            & D1  & 0.096     &\textbf {0.092}         & \textbf{}{0.092}     & 0.094          
\end{tblr}

\label{tab:7}
\end{table}

\begin{figure}[h]
    \centering
    %\begin{minipage}{0.6\textwidth}
        \centering
        \includegraphics[width=0.5\textwidth]{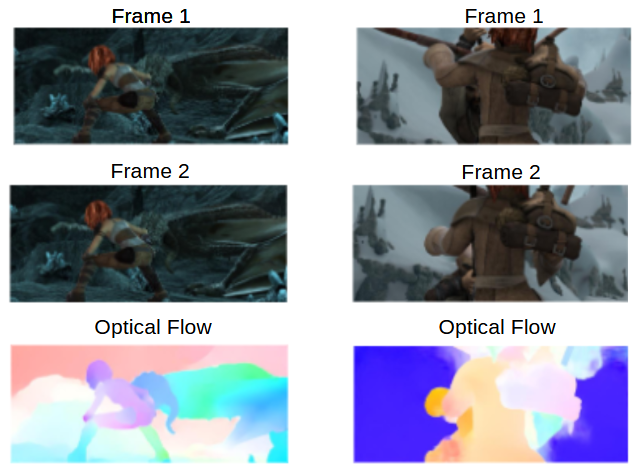}
        \caption{Visual representation of Optical Flow task on Sintel on two samples(left column and right column).}
        \label{fig:optical_flow}
   % \end{minipage}
\end{figure}

In Figure~\ref{fig:optical_flow}, the optical flow on the Sintel dataset demonstrates the model's ability to capture fine motion between frames with high precision, as evidenced by the smooth and well-defined flow fields.

\begin{table*}[]
\scriptsize
\caption{Disparity results across datasets (KITTI15, VKITTI, Sintel). D1 reported between 0-1}
\vspace{-2.5mm}
\centering
\begin{tblr}{
  width = \linewidth,
  colspec = {Q[130]Q[50]Q[50]Q[50]Q[50]Q[50]Q[50]Q[50]Q[50]},
  hline{1,2,3,4,5,6,7,8,9,10,11,12,13,14} = {-}{},
  vline{1,2,4,6,8,10,12} = {-}{},
  vline{3,5,7,9,11} = {2-13}{}
}
\textbf{Method} & \SetCell[c=2]{c}{\textbf{Kitti15}\cite{geiger2013vision}} && \SetCell[c=2]{c}{\textbf{Vkitti2}\cite{cabon2020virtual}} && \SetCell[c=2]{c}{\textbf{Sintel}\cite{butler2012naturalistic}} \\
& \textbf{EPE} & \textbf{D1} & \textbf{EPE} & \textbf{D1} & \textbf{EPE} & \textbf{D1} & \textbf{FPS} & \textbf{Memory}\\
\textbf{Unimatch} ~\cite{xu2023unifying}('23)  & 1.21  & 0.05  & 1.95  & 0.13 & 1.45  & \textbf{0.04} & 43.6 & 231.45\\
\textbf{IGEV} ~\cite{xu2023iterative}('23)     & 0.28  & 0.03  & 0.92&0.06 & \textbf{0.32}  & 0.12& 1.85&119.18 \\
\textbf{RAFT} ~\cite{lipson2021raft}('20)     & 1.08  & 0.05  & 0.92 & 0.06 & 0.45  & 0.13 & 2.82 &\textbf{102.41} \\
\textbf{Anynet} ~\cite{chen2022improvement}('18)   & 10.94 & 1.00  & 88.55 & 0.99  &  88.04 & 0.99 & 36 & 240.71\\
\textbf{Vision Mamba} ~\cite{zhu2024vision}('24)   & 1.38 & 0.07  & 1.14 & 0.06  &  11.53 & 0.24 & 52.53 & 334\\
\textbf{Mamba Vision} ~\cite{mambavision}('24)   & 0.71 & 0.02  & 1.04 & 0.06  &  1.43 & 0.06 & 52.26 & 224.85\\
\textbf{Mamba2}('24)  & 0.38  & 0.013  & 0.398  & 0.045  & 0.57  & 0.094 & 44.5&309.11 \\
\textbf{Mamba2 w/o attention}('24)  & 0.30  & 0.016  & 0.31  & 0.044  & 0.35  & 0.095 & 53.46 &286.71 \\
\textbf{VSSD}('24)  & 0.33  & 0.018  & 0.41  & 0.048  & 0.95  & 0.096 & 33.33 & 281.52 \\
\textbf{Proposed w/o Fusion}  & 0.28  & 0.012  & 0.281  & \textbf{0.043}  & 0.71  & 0.099 & \textbf{56.59} &297.29 \\
\textbf{Proposed}  & \textbf{0.27}  & \textbf{0.005}  & \textbf{0.18}  & 0.044  & 0.51  & 0.096 & 46.03 &324.85 \\
\end{tblr}
\vspace{1em}
\end{table*}
\label{tab_disparity_results}

%\begin{table}[!ht]
%\centering
%\caption{Ablation study for flow task on Sintel}
%\vspace{-1.5mm}
%\label{depth_results}
%\scriptsize
%\begin{tabular}{|l|r|r|r|r|r|r|r|r|}
%\hline
%Dataset &Metric &SA Inter &SA Cont &CA Inter &CA Cont &Both Inter &Both Cont \\
%\hline
%\multirow{6}{*}{Sintel (Clean)} &EPE &1 &2 &3 &4 &5 &6 \\
%\cline{2-8}&Matched &1 &2 &3 &4 &5 &6 \\
%\cline{2-8}&Unmatched &1 &2 &3 &4 &5 &6 \\
%\cline{2-8}&S0-10 &1 &2 &3 &4 &5 &6 \\
%\cline{2-8}&S10-40 &1 &2 &3 &4 &5 &6 \\
%\cline{2-8}&S40+ &1 &2 &3 &4 &5 &6 \\
%\hline
%\multirow{6}{*}{Sintel (Final)} &EPE &1 &2 &3 &4 &5 &6 \\
%\cline{2-8}&Matched &1 &2 &3 &4 &5 &6 \\
%\cline{2-8}&Unmatched &1 &2 &3 &4 &5 &6 \\
%\cline{2-8}&S0-10 &1 &2 &3 &4 &5 &6 \\
%\cline{2-8}&S10-40 &1 &2 &3 &4 &5 &6 \\
%\cline{2-8}&S40+ &1 &2 &3 &4 &5 &6 \\
%\hline
%\end{tabular}
%\end{table}

%-------------------------------------

\subsection{Disparity task}
%\subsubsection{Implementation details and Metrics}
%For stereo disparity, models are trained on the KITTI15, VKITTI2 and Sintel for 100k steps with batch size 6 and LR of 2e-4. We evaluate performance using common metrics like EPE, D1, FPS, and memory usage. EPE represents the average L1 distance between predicted and ground truth disparity, whereas D1 indicates the percentage of outliers.

\subsubsection{Results and Analysis} 
The proposed model demonstrates clear improvements in disparity estimation across multiple datasets, particularly in reducing outliers and enhancing efficiency (see Table~\ref{tab_disparity_results}).

On the KITTI15 dataset, it achieves an {EPE of 0.27} and {D1 of 0.005}, significantly outperforming RAFT and Unimatch in outlier reduction. While IGEV attains a similar EPE, its higher D1 (0.37) underscores the importance of balancing accuracy with consistency, which the proposed model handles effectively. On VKITTI, the model records an {EPE of 0.18} and {D1 of 0.044}, again outperforming RAFT and IGEV in outlier percentage, highlighting its robustness in handling complex synthetic datasets. The large error reduction achieved through cross-attention mechanisms further validates its reliability in difficult scenarios. Similarly, on the Sintel dataset, the model performs comparably to better to Unimatch with an {EPE of 0.51} and {D1 of 0.096}, maintaining a strong balance between accuracy and outlier minimization, even in dynamic scenes. In terms of efficiency, the proposed model delivers superior performance compared to RAFT, Anynet, and IGEV is slightly better than Unimatch (FPS of 43.6), achieving {46.03 FPS}, while maintaining a competitive memory footprint of {324.85 MB}, considering the significant improvements in EPE and D1. 
In comparison to VSSD, Mamba2, Mamba2 w/o attention and proposed w/o fusion, VisionMamba and MambaVision in the Kitti15 and Vkitti2 datasets out method outperformed. Moreover, the results from Table~\ref{tab_disparity_results} clearly demonstrate that the model is highly suitable for real-time applications, combining speed and scalability with robust accuracy. 

\begin{figure}[h]
    \centering
    %\begin{minipage}{0.7\textwidth}
        \centering
        \includegraphics[width=0.5\textwidth]{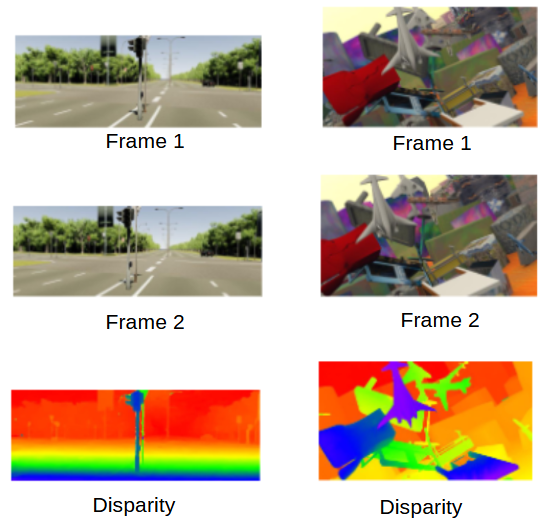}
        \caption{Visual representation of Disparity task on Vkitti2 and Sintel.}
        \label{fig:disparity_map}
   % \end{minipage}
\end{figure}

Finally, Figure~\ref{fig:disparity_map} presents the disparity maps generated from the Vkitti2 and Sintel datasets. The vivid disparity maps reveal our model's robustness in estimating pixel-wise disparities in challenging scenes, contributing to high-quality 3D scene reconstruction.

\subsubsection{Ablation on disparity task}
The ablation study in Table \ref{disp_abalation} highlights the performance variations between different attention mechanisms (cross-attention, self-attention, a combination of both and no attention) across three disparity datasets: KITTI15, VKITTI, and Sintel. Overall, the combination of self and cross-attention consistently achieves better accuracy, particularly in the VKitti2 dataset, where it outperforms other configurations in reducing End-Point Error (EPE) and D1. On Kitti15, the combined self and cross-attention show slightly better performance in EPE and maintain relatively low D1 errors, highlighting the robustness of the cross-attention combined with self-attention. However, on the Sintel dataset, all configurations exhibit higher EPEs, indicating that the proposed architecture may require further refinement for complex, dynamic environments. The results confirm that cross-attention combined with self-attention provides a well-balanced solution for disparity estimation, minimizing both error and outliers across diverse datasets.
\\

\subsection{Flow to Disparity}
The results of transferring pre-trained weights from optical flow tasks to disparity estimation demonstrate compelling findings, as outlined in Table \ref{ftod}. Specifically, when training the disparity estimation model using pre-trained weights derived from an optical flow task, we observe superior performance compared to training from scratch on the KITTI15 dataset. The model initialized with pre-trained weights achieves an End-Point Error (EPE) of 0.31 and a D1 score of 0.005, outperforming the model trained from scratch, which reaches an EPE of 0.39 and a D1 score of 0.006. This indicates that initializing disparity estimation training with optical flow-pretrained weights leads to a beneficial effect, yielding better accuracy metrics on this dataset. This also proves that the proposed model can produce a unified model for the dense perfection task of flow and disparity.

\begin{table}[!ht]
\tiny
\caption{Result on cross task transfer from flow to disparity on KITTI15.}
\resizebox{\columnwidth}{!}{%
\begin{tabular}{|c|c|c|c|c|c|}
\hline
Method &                EPE & D1  \\ \hline
Flow to Disparity on Proposed&\textbf{0.31}&\textbf{0.005} \\ \hline
Proposed from scratch &              0.39 & 0.006  \\ \hline
% Proposed w/o Fusion &   0.41 & 0.008  \\ \hline
% Mamba2 &                 0.41 & 0.007   \\ \hline
% Mamba2 w/o Attention &  0.42& 0.008   \\ \hline
% VSSD\cite{shi2024vssdvisionmambanoncausal} &                  0.46 & 0.01  \\ \hline
\end{tabular}%
}
\label{ftod}
\end{table}
% \begin{table}[!ht]
% \caption{Ablation study on cross task transfer from flow to disparity on KITTI15.}
% \resizebox{\columnwidth}{!}{%
% \begin{tabular}{|c|c|c|c|c|c|}
% \hline
% Method &                EPE & D1  \\ \hline
% Flow to Disparity on Proposed&\textbf{0.31}&\textbf{0.005} \\ \hline
% Proposed from scratch &              0.39 & 0.006  \\ \hline
% % Proposed w/o Fusion &   0.41 & 0.008  \\ \hline
% % Mamba2 &                 0.41 & 0.007   \\ \hline
% % Mamba2 w/o Attention &  0.42& 0.008   \\ \hline
% % VSSD\cite{shi2024vssdvisionmambanoncausal} &                  0.46 & 0.01  \\ \hline
% \end{tabular}%
% }
% \label{ftod}
% \end{table}

\section{Conclusions}
\label{ref:conclusion}

This work introduces a novel Mamba block DenVisCoM, as well as a novel unified hybrid architecture for accurate and real-time estimation of optical flow and disparity estimation. To nurture the dense correspondence of the tasks, the features from the pair of input images are fused and passed through the sequence transformation pipeline \textit{ ie} the Scan branch of the Mamba block, which consists of the SSM component. The proposed hybrid architecture uses the DenVisCoM and Transformer-based self- and cross-attention block for unified motion and 3D dense perception tasks. To conclude, experimental results and analysis showcase that the proposed model was able to tackle speed, accuracy, and memory gaps better than state-of-the-art techniques.

\bibliographystyle{IEEEtran}
\bibliography{icra/main}

\end{document}